# A Diverse Corpus for Evaluating and Developing English Math Word Problem Solvers

**Shen-Yun Miao, Chao-Chun Liang, and Keh-Yih Su**
Institute of Information Science, Academia Sinica, Taiwan
miao.iisas@gmail.com, {ccliang, kysu}@iis.sinica.edu.tw

## Abstract

We present *ASDiv* (**A**cademia **S**inica **Di**verse MWP Dataset), a diverse (in terms of both language patterns and problem types) English *math word problem* (MWP) corpus for evaluating the capability of various MWP solvers. Existing MWP corpora for studying AI progress remain limited either in language usage patterns or in problem types. We thus present a new English MWP corpus with 2,305 MWPs that cover more text patterns and most problem types taught in elementary school. Each MWP is annotated with its *problem type* and *grade level* (for indicating the level of difficulty). Furthermore, we propose a metric to measure the *lexicon usage diversity* of a given MWP corpus, and demonstrate that *ASDiv* is more diverse than existing corpora. Experiments show that our proposed corpus reflects the true capability of MWP solvers more faithfully.

## 1 Introduction

Human math/science tests have been considered more suitable for evaluating AI progress than the Turing test (Clark and Etzioni, 2016). Among them, *math word problems* (MWPs) are frequently chosen to study natural language understanding and simulate human problem solving (Bakman, 2007; Mukherjee and Garain, 2008; Liang et al., 2016), because the answer is not a span within the given problem text that can be directly extracted. Table 1 shows a typical example of MWP, which consists of a few sentences that involve quantities.

Current MWP corpora can be classified into four categories: (1) the *Number Word Problem* corpus (Shi et al., 2015), which contains number word problems only; (2) the *Arithmetic Word Problem* corpora (Hosseini et al., 2014; Roy et al., 2015), which involve the four basic arithmetic operations

| Math Word Problem |
|---|
| A sandwich is priced at $0.75. A cup of pudding is priced at $0.25. Tim bought 2 sandwiches and 4 cups of pudding. How much money should Tim pay? |
| **Solution:** 0.75 x 2 + 0.25 x 4 = 2.5 |

Table 1: A math word problem

(*addition*, *subtraction*, *multiplication* and *division*) with either single-step or multi-step operations; (3) the *Algebraic Word Problem* corpora (Kushman et al., 2014; Koncel-Kedziorski et al., 2015; Roy and Roth, 2017; Upadhyay & Chang, 2015; Wang et al., 2017), which focus on algebraic MWPs; and (4) the *Mixed-type MWP* corpora (Huang et al., 2016, Ling et al., 2017, Amini et al., 2019), which are large-scale collections of either daily algebra or GRE/GMAT examination MWPs. Table 2 is a comparison of existing English MWP corpora.

However, these existing corpora are either limited in terms of the diversity of the associated *problem types* (as well as *lexicon usage patterns*), or lacking information such as *difficulty levels*. For example, categories (1), (2), and (3) collect only certain types of MWPs. On the other hand, although large-scale mixed-type MWP corpora contain more problem types, the annotated answers or formulas are sometimes inconsistent, and the corresponding difficulty level is usually not provided.

Furthermore, low-diversity corpora are typically characterized by highly similar problems, which usually yields over-optimistic results (Huang et al., 2016) (as the answer frequently can be simply obtained from the existing equation template associated with the most similar MWP in the training-set). Roy and Roth (2017) shown significantly lowered performance if highly similar MWPs are removed. Therefore, *dataset diversity* is more critical than the *dataset size* for accurately judging the true capability of an MWP solver.

We thus present *ASDiv* (**A**cademia **S**inica **Di**verse MWP Dataset), a new MWP corpus that contains diverse lexicon patterns with wide problem type coverage. Each problem provides consistent equations and answers. It is further annotated with the corresponding *problem type* and *grade level*, which can be used to test the capability of a system and to indicate the difficulty level of a problem, respectively. The diverse lexicon patterns can be used to assess whether an MWP solver obtains answers by understanding the meaning of the problem text, or simply by finding an existing MWP with similar patterns (Huang et al., 2016). Problem type diverseness is crucial for evaluating whether a system is competitive with humans when solving MWPs of various categories. Besides, to assess text diversity, we propose a *lexicon usage diversity* metric to measure the diversity of an MWP corpus.

| Corpus | MWP category | Annotation | Size |
| --- | --- | --- | --- |
| Dolphin1878 (Shi et al., 2015) | Number-word problems | Equation/answer | 1,878 |
| AI2 (Hosseini et al., 2014) | Arithmetic word problems | Equation/answer | 395 |
| IL (Roy et al., 2015) | Arithmetic word problems | Equation/answer | 562 |
| AllArith (Roy and Roth, 2017) | Arithmetic word problems | Equation/answer | 831 |
| KAZB (Kushman et al., 2014) | Algebraic word problems | Equation/answer | 514 |
| ALGES (Koncel-Kedziorski et al., 2015) | Algebraic word problems | Equation/answer | 508 |
| DRAW (Upadhyay & Chang, 2015) | Algebraic word problems | Equation/answer/template | 1,000 |
| Dolphin18K (Huang et al., 2016) | Arithmetic/algebraic + domain knowledge problems | Equation/answer | 18K |
| AQuA (Ling et al., 2017) | Arithmetic/algebraic + domain knowledge problems | Rationale/answer (multi-choice problems) | 100K |
| MathQA (Amini et al., 2019) | Arithmetic/algebraic + domain knowledge problems | Decomposed-linear formula/answer (multi-choice problems) | 37K |
| ASDiv | Arithmetic/algebraic + domain knowledge problems | Equation/answer + grade-level/problem-type | 2,305 |

Table 2: Comparison of different English MWP corpora

This paper makes the following contributions: (1) We construct a diverse (in terms of lexicon usage), wide-coverage (in problem type), and publicly available[1] MWP corpus, with annotations that can be used to assess the capability of different systems. (2) We propose a *lexicon usage diversity metric* to measure the diversity of an MWP corpus and use it to evaluate existing corpora. (3) We show that the real performance of state-of-the-art (SOTA) systems is still far behind human performance if evaluated on a corpus that mimics a real human test.

## 2 Problem Type

A *problem type* (*PT*) indicates a crucial math operation pattern for solving an MWP. As MWPs of the same problem type share a similar pattern (in language usages, logic representation, or inferences), they thus indicate stereotypical math operation patterns that could be adopted to solve an MWP (Liang et al., 2018). In *ASDiv*, each MWP is annotated with a specific *PT* taught at elementary schools. Some examples of selected *PTs* are shown in Table 5 (Appendix). Currently, we provide 24 different common *PTs* and classify them into three main categories according to the operations involved and illustrated below. These *PTs* are usually specified in math textbooks and mostly covered in elementary schools.

**Basic arithmetic operations:** This category includes: *Addition*, *Subtraction*, *Difference*, *Multiplication*, three different *Divisions* (i.e., *common-division*, *floor-division*, and *ceil-division*), *Sum*, *Surplus*, *Number-Operation*, three different ***T**ime-**V**ariant-**Q**uantities* (TVQ), and *Multi-step*. The first seven types are self-explanatory. *Number-Operation* indicates that the problem description consists mainly of numbers and their relations. $TVQ^2$ denotes an entity-state related variable (e.g., *initial/current/final-state* and *change*) whose value is updated sequentially according to a sequence of events described in an MWP. Last, in a *Multi-step* problem, the answer is obtained from multiple arithmetic operations.

**Aggregative operations:** This category includes: (1) *Comparison*, (2) *Set-Operation*, (3) *Ra*

---

[1] The corpus can be found at: *https://github.com/chaochun/nlu-asdiv-dataset*.

[2] E.g., "*the number of apples that Jack has*" is a *TVQ* in "*Jack had 5 apples; he then ate 3 of them. How many apples does Jack has now?*".

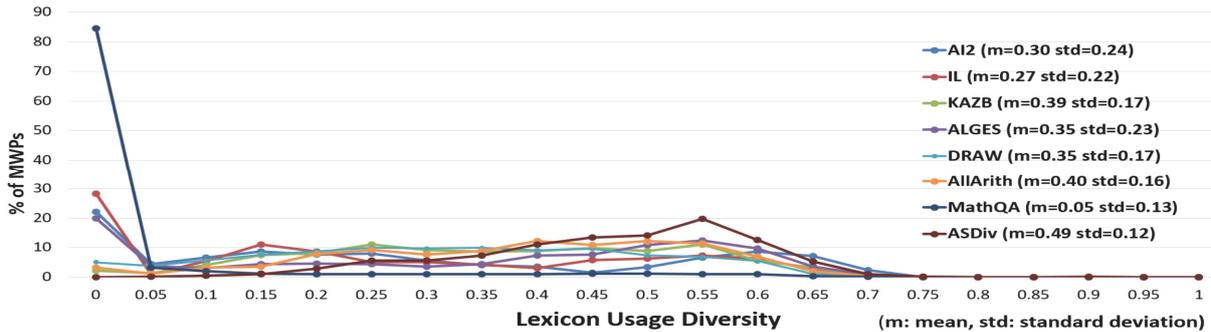

Figure 1: Lexicon usage diversity of various corpora.

*tio,* (4) *Number-Pattern*, (5) *Algebra-1*, and (6) *Algebra-2*. The first three types are self-explanatory. *Number-Pattern* refers to the problems which involve deducing a pattern from a sequence of integers (Table 5 (Appendix) shows an example). *Algebra-1* and *Algebra-2* are algebraic problems with one and two unknown variables, respectively.

**Additional domain knowledge required:** This category includes *Greatest Common Divisor*, *Least Common Multiple*, *Geometry*, and *UnitTrans*. Additional geometric knowledge (e.g., *area = length * width*) is required in *Geometry* problems. *UnitTrans* means that the answer is obtained via conversion to the metric system (e.g., converting '*miles*' to '*kilometers*').

## 3 ASDiv Math Word Problem Corpus

This corpus was designed based on the following guidelines: (1) The corpus should be as diverse as possible in terms of *lexicon usage* so that the answer is less likely to be obtained via mechanical/statistical pattern matching without understanding the content. (2) The corpus should cover most *PTs* found in primary school so that it can approximate real human tests. (3) The corpus should be annotated with sufficient information so that it can be used not only to assess the capability of various systems but also to facilitate system development.

### 3.1 Corpus Diversity Metrics

We first propose a *lexicon usage diversity* metric, in terms of BLEU (Papineni et al., 2002), to measure the degree of diversity of a given corpus. This metric is from 0 to 1; higher value indicates the corpus is more diverse. We first use Stanford *CoreNLP* (Manning et al., 2014) to tokenize and tag POSs, and then use *NLTK* (Bird et al., 2004) to lemmatize each token. Furthermore, we normalize the original sentences with: (1) stop word removal; and (2) named entity and quantity normalization, which replace the associated person names and quantity values with meta symbols in an MWP (i.e., two MWPs are regarded as identical if they differ only in names or quantity-values). This thus places the focus on essential words that matter in the MWP. The obtained sequence is then used to measure the lexicon usage diversity specified below.

Let $P = \{P_1, P_2, ..., P_M\}$ be a specific set of MWPs in a given corpus with the same *PT*, where $P_i$ is the *i*-th MWP in $P$. For a given $P_i$, we define its *lexicon usage diversity* (LD) of $P_i$ as

$$LD_i = 1 - \max_{j, j \neq i} \frac{BLEU(P_i, P_j) + BLEU(P_j, P_i)}{2},$$

where $BLEU(P_i, P_j)$ is the BLEU score between $P_i$ and $P_j$ ( $j \neq i$, $j = \{1, 2, ..., M\}$ ). We measure the BLEU score *bi-directionally* with n-grams up to *n=4*. This measure is mainly used to identify repeated usage of lexicon and phrases; *n=4* suffices for this case. $LD_i$ evaluates the lexicon diversity between $P_i$ and all $P_j$ ( $i \neq j$ ). Furthermore, the *mean* of all $LD_i$ (under the same corpus) can be used to indicate the *corpus lexicon diversity* (CLD). Adding a new MWP with a low LD to an existing corpus introduces little new information to the corpus; thus, it should be either discarded or revised. This diversity metric can help the corpus constructor to decide whether an MWP can be directly adopted or not.

### 3.2 Challenges in Constructing a Large-Scale MWP Dataset

Since MathQA is the second-largest dataset in Table 2 (with 37K MWPs), and is cleaner (Amini et al., 2019) than the largest one (AQuA), we first evaluate it with the above LD measurement. Figure 1 shows that its CLD is only 0.05.

To understand the reason for the low diversity of MathQA (LD = 0 for 85% of the MathQA MWPs), we investigated this dataset. We observed that MathQA includes various MWP subsets, each of which shares the same sentence pattern among its members. Figure 1 clearly shows its skewed distribution. Figure 3 (Appendix) shows a subset of which all 105 members share the same sentence pattern.

Since most MWP solvers can only solve arithmetic MWPs, we further selected its arithmetic[3] subset, generated their corresponding equations according to the annotated formulas, and then solved the equations using the SymPy[4] package. Afterwards, we verified the consistency between the answer obtained from the annotated formula and the labeled answer. The results show that the annotated formulas of 27% of the problems do not match their labeled answers.

We randomly inspected 30 inconsistent MWPs and classified them into three error-types: (1) Incorrect formula (67%), for which the annotated formula cannot be used to solve the given MWP; (2) problematic description (23%), for which the description text is either incomplete or problematic; (3) valueless answer (10%), for which the given answer is either wrong or inappropriate. Table 6 (Appendix) illustrates examples of each error-type.

Although building a large corpus via crowdsourcing is a tempting approach, it can result in a poor-quality corpus if the annotation procedure is not well controlled. We believe the *quality* of the dataset is more important than its *size*, if they cannot be achieved simultaneously.

### 3.3 Corpus Construction

To account for the problems observed in MathQA, we first collected MWPs from 28 websites and then either pruned the problem or revised the text if it was highly similar to any existing ones (according to the proposed *lexicon usage diversity* metric). This yielded a total of 2,305 MWPs.

Next, we hired one master-degree research assistant with a background in automatic MWP solving to annotate the problem type, equation, answer, and grade level manually for each MWP. If annotations were provided with the original MWP (22.6% of the source MWPs included equations and answers; 52% had answers only; 63.5% included grade-level information), we used it directly; otherwise, we annotated them manually[5].

Since MWPs are usually clearly specified (with a sure answer), there is no ambiguous interpretation once the answer is given. Therefore, as opposed to other corpora in which annotations (mostly linguistic attributes) are mainly based on human *subjective* judgment, the MWP answer/equation annotation is more *objective* and must be consistent. As a result, human *carefulness*, instead of human *agreement*, is a more critical issue in this task. Since an incorrect math expression usually yields an incorrect answer, we used a program to automatically verify the consistency between the annotated equations and the answers. Inconsistent MWPs were re-annotated and checked again. Afterwards, we randomly selected 480 samples (20 samples per problem type) to verify the final annotation correctness. All those samples were correct, which confirms our above assertion.

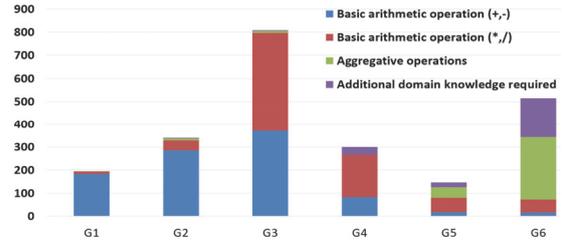

Figure 2: Distribution of PT categories (G1~G6)

Figure 2 shows the distribution of different problem categories in six grade levels in elementary school. Most arithmetic operations appear in grade levels 1 to 4, which means students learn basic arithmetic operations in this stage. We further separate *Addition/Subtraction* from *Multiplication/Division* to highlight that they are in different difficulty levels for students. Figure 2 also indicates Multiplication/Division is more emphasized in grade 3 and 4. In grades 5 and 6, improved math skills enable students to solve difficult MWPs that require more aggregative operations and additional domain knowledge. Thus, the grade level is a useful indicator of difficulty and can be employed to evaluate the capability of MWP solving systems.

### 3.4 LD Distributions of Various Corpora

We compare the diversity among various MWPs of the same *PT* (for those corpora without annotated PT Category, diversity is measured over the whole corpus). Lastly, we generate the associated *LD* distributions (uniformly quantized into 20 intervals between 0 and 1) and calculate the *corpus lexicon*

---

[3] Associated formula involves only arithmetic operations.
[4] https://www.sympy.org/en/index.html.
[5] We annotated the grade level according to the following grade classification reference: (1) Achieve Inc. 2008. Out of many, one: Toward rigorous common core standards from the ground up. Washington, DC; Common Core State; (2) Standards Initiative. 2010. Common Core State Standards for mathematics.

|   | MathQA-C (CLD=0.08) | ASDiv-A (CLD=0.50) | ASDiv (CLD=0.49) |
|---|---|---|---|
| L | - | 0.68 | 0.36 |
| U | - | 0.78 | 0.37 |
| G | 0.86 | 0.68[#] | 0.36[#] |

Table 3: Accuracies for different systems (CLD denotes the corpus lexicon diversity; L, U and G denote the *LCA++*, *UnitDep,* and *GTS* systems respectively. '-' denotes failure on this corpus; # indicates performance is significantly lower than "-C" with p<0.01.

|   | G1 | G2 | G3 | G4 | G5 | G6 |
|---|---|---|---|---|---|---|
| L | 0.53 | 0.64 | 0.49 | 0.35 | 0.03 | 0.01 |
| U | 0.55 | 0.65 | 0.51 | 0.34 | 0.03 | 0.01 |
| G | 0.64 | 0.60 | 0.47 | 0.34 | 0.07 | 0.01 |

Table 4: Performance of various grade levels on the ASDiv. L/U/G are the same as that in Table 3.

*diversity* (CLD, Section 3.1) on corpora frequently adopted for comparing various systems: (1) AI2, (2) IL, (3) KAZB, (4) ALGES, (5) DRAW, (6) AllArith, and (7) MathQA.

Figure 1 shows the distributions of CLD for various corpora: there are about 85%, 28%, 22% and 20% identical MWPs (these numbers are the percentages of MWPs with $LD_i=0$ w.r.t. each dataset) in MathQA, IL, AI2 and ALGES corpora respectively, whereas ASDiv contains none. We also evaluate *syntactic pattern diversity* (in terms of *POS n-gram*) and the diversity between MWPs in the training set and the test set. Both yield similar trends, too (details are given in the Appendix).

## 4 Experiments

To study the correlation between CLD and system performance, we selected three SOTA MWP solvers to conduct the experiments: two based on statistical models, *LCA++* (Roy and Roth, 2015) and *UnitDep* (Roy and Roth, 2017); and one using a neural network which adopts two-layer gate-feedforward networks for a Goal-driven Tree-structured approach (*GTS*) (Xie et al., 2019).

Since the selected MWP solvers solve only arithmetic MWPs, we first collected 4,117 MWPs from MathQA to construct a subset that its associated formulas satisfy the following two conditions: (1) they involve only arithmetic operations; and (2) they contain neither external constants (which would necessitate external domain knowledge to solve the problem and is out of the scope of this work) nor reused operands (which rarely occur and would complicate the solution procedure). We filtered out inconsistent problems (specified in Section 3.2) and termed the remaining 3,000 MWPs as

*MathQA-C* dataset (-C for *consistent*) to evaluate the performance. Similarly, we extracted a subset of 1,218 MWPs that involve only arithmetic operations (and also satisfy the constraints mentioned above) from ASDiv, and termed this the *ASDiv-A* dataset (-A for *arithmetic*). The CLDs for MathQA-C and ASDiv-A were found to be 0.08 and 0.50, respectively. Also, LD = 0 for 82% of the MathQA-C MWPs.

Afterwards, we tested the solvers against three MWP corpora: MathQA-C, ASDiv-A, and ASDiv. MathQA-C is reported with 5-fold cross-validation accuracy. For ASDiv-A and ASDiv, we randomly split the MWPs of each *PT* into five nearly equally-sized subsets, and report the 5-fold cross-validation accuracy. For GTS system, we repeated the experiment 5 times and obtained the averaged answer accuracy.

Table 3 compares the answer accuracies of various systems. We observe that the overall performance is only around 36% on ASDiv, which shows that the performance of the current SOTA systems still is not competitive with human performance, and that CLD is correlated with the system performance (i.e., lower diversity implies higher performance) and is a useful metric to evaluate existing corpora. Table 4 further shows the accuracy of different grade levels on ASDiv: the performance of grades 5 and 6 are significantly lower than the performance of grade 1 to 4. As accuracy is highly correlated with the grade level, the grade level is a useful index for indicating the difficulty of MWPs.

## 5 Conclusion and Future Work

We present an MWP corpus which not only is highly diverse in terms of lexicon usage but also covers most problem types taught in elementary school. Each MWP is annotated with the corresponding problem type, equation, and grade level, which are useful for machine learning and assessing the difficulty level of each MWP. We also propose a metric to measure the diversity of lexicon usage of a given corpus. In terms of this metric, we show that in comparison with those corpora widely adopted to compare systems, ours is more suitable for assessing the real performance of an MWP solver. Last, we conduct experiments to show that a low-diverse MWP corpora will exaggerate the true performance of SOTA systems (we are still far behind human-level performance), and that grade level is a useful index for indicating the difficulty of an MWP.

# Appendix

## Appendix A: Examples of a few Selected Problem Types

Table 5 shows examples of selected types in "Basic arithmetic operations", "Aggregative operations", and "Additional domain knowledge required" categories.

| Problem type | Examples |
|---|---|
| **Basic arithmetic operations** | |
| Number-Operation | I have 3 hundreds, 8 tens, and 3 ones. What number am I? |
| TVQ-Initial | Tim's cat had kittens. He gave 3 to Mary and 6 to Sara. He now has 9 kittens. How many kittens did he have to **start with**? |
| TVQ-Change | For the school bake sale, Wendy made pastries. She baked 41 cupcakes and 31 cookies. After the sale, she had 32 to take back home. How many pastries did she sell? |
| TVQ-Final | Melanie had 7 dimes in her bank. Her dad gave her 8 dimes and her mother gave her 4 dimes. How many dimes does Melanie **have now**? |
| Multi-Step | They served a total of 179 adults and 141 children, if 156 of all the people they served are male, how many are female? (combination of multiple operations) *(Equation: x = (179+141)-156)* |
| **Aggregative operations** | |
| Algrbra-1 | Maddie, Luisa, and Amy counted their books. Maddie had 15 books. Luisa had 18 books. Together, Amy and Luisa had 9 more books than Maddie. How many books did Amy have? *(Equation: (x+18)-9 = 15, where x: "money of Amy")* |
| Algrbra-2 | The cost of a private pilot course is $1,275. The flight portion costs $625 more than the ground school portion. What is the cost of each? *(Equation: x+y = 1275; x = y+625, where x:The cost of flight portion; y:The cost of ground school portion)* |
| Comparison | A bookstore was selling 2 books for $15.86. You could buy 7 books for $55.93 online. Which place has a lower unit price? |
| Set-Operation | Sarah's team played 8 games of basketball. During the 8 games her team's scores were 69, 68, 70, 61, 74, 62, 65 and 74. What is the mean of the scores? |
| Ratio | There are 43 empty seats and 7 occupied seats on an airplane. What is the ratio of the number of occupied seats to the total number of seats? |
| Number-Pattern | The Wholesome Bakery baked 5 loaves of bread on Wednesday, 7 loaves of bread on Thursday, 10 loaves of bread on Friday, 14 loaves of bread on Saturday, and 19 loaves of bread on Sunday. If this pattern continues, how many loaves of bread will they bake on Monday? |
| **Additional domain knowledge required** | |
| G.C.D. | A teacher is to arrange 60 boys and 72 girls in rows. He wishes to arrange them in such a way that only boys or girls will be there in a row. Find the greatest number of students that could be arranged in a row. |
| L.C.M. | On a track for remote-controlled racing cars, racing car A completes the track in 28 seconds, while racing car B completes it in 24 seconds. If they both start at the same time, after how many seconds will they be side by side again? |
| UnitTrans | Mrs. Hilt will buy a new pair of shoes in 11 days. How many minutes must she wait before she can buy her new pair of shoes? |

Table 5: Examples of selected problem types

# Appendix B: Problematic MWPs in MathQA

Table 6 shows examples of inconsistent MWPs in MathQA, and Figure 3 shows examples using the same sentence pattern.

| Error types | Examples | Annotated Formula | Labeled Answer |
|---|---|---|---|
| Incorrect formula (67%) | Problem-1 "What is the 25 th digit to the right of the decimal point in the decimal form of 6 / 11 ?" | Annotation Formula: **divide**(n1,n2) <br> Desired Formula: Not available[6] | 6 |
| | Problem-2 "In two triangles, the ratio of the areas is 4 : 3 and that of their heights is 3 : 4. find the ratio of their bases." | Annotation Formula: multiply(n0,n0)\|multiply(n1,n1)\|**divide**(#0,#1) <br> Desired Formula: Not available[7] | 16:9 |
| Problematic description (23%) | Problem-3 "The lcm of two numbers is 495 and their hcf is 5. if the sum of the numbers is **10**, then their difference is" <br> Desired text: "… sum of the numbers is **100**" | Annotation Formula: multiply(n0,n1)\| divide(#0,n2) | 10 |
| Valueless answer (10%) | Problem-4 "9886 + x = 13200, then x is ?" <br> Options: a) 3327, b)3237, c)3337, d) 2337, e)none of these | Annotation Formula: subtract(n1,n0) | Annotation Option (e): none of these <br> Desired Answer: 3414 |

Table 6: Some examples of inconsistent answers on MathQA.

*A train running at the speed of **100** km/hr crosses a pole in **18** seconds. What is the length of the train ?*
*A train running at the speed of **100** km/hr crosses a pole in **9** seconds. What is the length of the train ?*
*A train running at the speed of **108** km/hr crosses a pole in **7** seconds. Find the length of the train ?*
*A train running at the speed of **110** km/hr crosses a pole in **9** seconds. What is the length of the train ?*
*A train running at the speed of **120** km/hr crosses a pole in **12** seconds. What is the length of the train ?*
*A train running at the speed of **120** km/hr crosses a pole in **18** seconds. What is the length of the train ?*
*A train running at the speed of **120** km/hr crosses a pole in **9** seconds. Find the length of the train ?*
*A train running at the speed of **126** km/hr crosses a pole in **9** seconds. Find the length of the train ?*
*A train running at the speed of **142** km/hr crosses a pole in **12** seconds. Find the length of the train ?*
*A train running at the speed of **162** km/hr crosses a pole in **9** seconds. Find the length of the train ?*
*A train running at the speed of **180** km/hr crosses a pole in **18** seconds. What is the length of the train ?*
*A train running at the speed of **180** km/hr crosses a pole in **36** seconds. What is the length of the train ?*
*A train running at the speed of **180** km/hr crosses a pole in **7** seconds. What is the length of the train ?*
*A train running at the speed of **180** km/hr crosses a pole in **8** seconds. Find the length of the train ?*
*A train running at the speed of **180** km/hr crosses a pole in **9** seconds. Find the length of the train ?*

Figure 3: Examples in MathQA with the same sentence pattern (after sentence normalization). These MWPs all share the same sentence pattern, "*a train running at the speed of [NUM1] km/hr crosses a pole in [NUM2] sec. [what is | find] the length of the train?*".

---

[6] There is no proper operation on the MathQA operation set.

[7] There is no proper operation on the MathQA operation set.

# Appendix C: Additional Experiments for Corpus Diversity Metrics

We also provide a *syntactic pattern diversity* to measure the syntactic diversity of an MWP. Let $P = \{P_1, P_2, ..., P_M\}$ be a specific set of MWPs in a given corpus with the same *problem type*, where $P_i$ is the $i$-th MWP in $P$, and $T_i$ is the corresponding POS sequence of $P_i$. For example, "NNP VBZ CD NNS" is the POS tagging sequence for "Mary has 5 books.". For a given $P_i$, we define its *syntactic pattern diversity* ($SD_i$) as

$$SD_i = 1 - \max_{j, j \neq i} \frac{BLEU(T_i, T_j) + BLEU(T_j, T_i)}{2},$$

where $BLEU(T_i, T_j)$ is measured between $T_i$ and $T_j$ ( $j \neq i, j = \{1, 2, ..., M\}$ ). Figure 4 shows that there are 87%, 54%, 46% and 33% identical syntactic patterns (these numbers are the percentages of MWPs with $SD_i = 0$ w.r.t. each dataset) in the MathQA, IL, AI2, and ALGES corpora, respectively, while ASDiv only has 4%. This shows that our corpus is also more diverse in terms of syntactic patterns.

We also measure the diversity between the test-set and the training-set, as the similarity between them is a critical factor[8] for causing exaggerated performance. The *LD* and *SD* metrics between the test-set and the training-set can be obtained by modifying the previous formulas to

$$LD_i = 1 - \max_j \frac{BLEU(P_i, P_j) + BLEU(P_j, P_i)}{2},$$
$$P_i \in DS_{test} \text{ and } P_j \in DS_{train}$$

$$SD_i = 1 - \max_j \frac{BLEU(T_i, T_j) + BLEU(T_j, T_i)}{2},$$
$$T_i \in DS_{test} \text{ and } T_j \in DS_{train}$$

where $DS_{test}$ and $DS_{train}$ are all the MWPs in the test set and the training set, respectively. For a given problem $P_i$ from the test set, $LD_i$ and $SD_i$ denote the lexicon-pattern and syntactic-pattern diversity between $P_i$ and all the problems $P_j$ (in the training set), respectively. If $P_i$ has a low diversity index, it can be easily solved via a training set MWP with similar patterns.

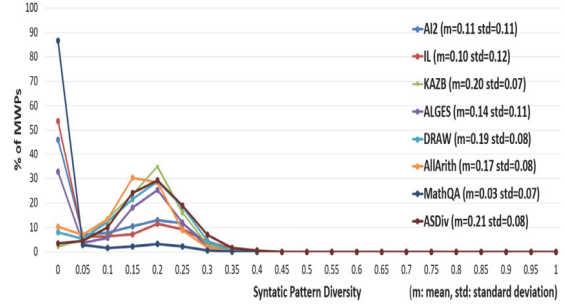

Figure 4: Syntactic pattern diversity of various corpora

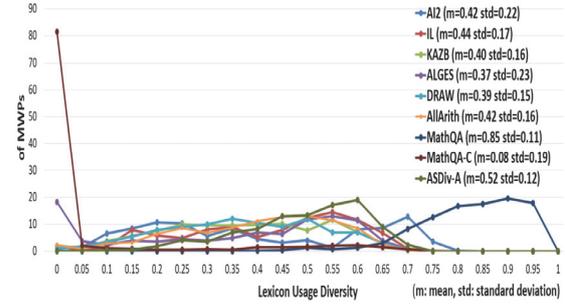

Figure 5: Lexicon usage diversity of various corpora: *test-set* versus *training-set*

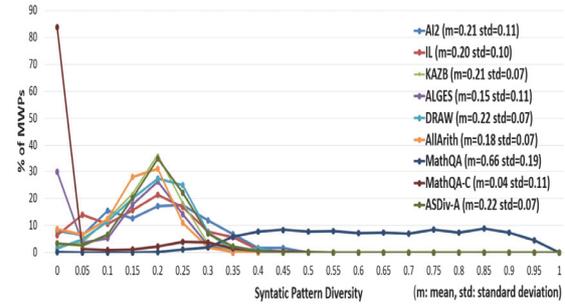

Figure 6: Syntactic pattern diversity of various corpora: *test-set* versus *training-set*

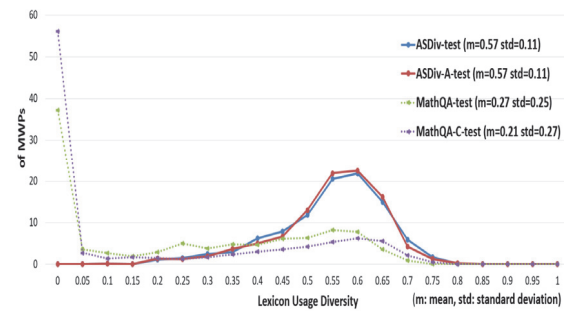

Figure 7: Lexicon usage diversity measured only within the test-set for various corpora. Here, the suffix "-test" denotes that it is measured only within the test-set.

---

[8] However, we also need another critical factor - the *CLD* value within the test-set - to assess the true value of a given corpus. Suppose we take an MWP with a high LD value from the existing training-set, and then duplicate it 300 times to create a new test-set. This would still result in a high *CLD* value between the training-set and this new test-set. However, this setting would be meaningless in terms of assessing the true performance of MWP solvers. Likewise, we could also enlarge the training-set by duplicating MWPs without affecting the *CLD* value against the test-set. However, it would be also meaningless as no new information would be provided.

For AI2, IL, KAZB, ALGES, AllArith, DRAW, and MathQA, we follow their own either n-fold cross-validation setting or the train/dev/test splits originally specified in each dataset. For MathQA-C and ASDiv-A, we follow the same n-fold cross-validation setting specified in Section 4. Figure 5 shows the LD between the test-set and the training-set for various corpora.

Also, though not directly shown in this figure, the lexicon diversity indices of 48% of the MWPs in our ASDiv-A are larger than or equal to 0.5. In contrast, they are 35%, 34%, 29%, 23%, 23%, 18% and 7% in AI2, IL, ALGES, KAZB, AllArith, DRAW, and MathQA-C respectively. These statistics suggest that our corpus is more diverse (and thus more difficult) than other well-known MWP corpora. Figure 6 shows the SD between the test-set and the training-set for different corpora. Again, this shows that our corpus also contains more diverse POS sequences.

Last, in Figure 5, MathQA actually possesses the highest CLD between its official test-set and training-set: 0.85, surprisingly higher than that of all other corpora (e.g., they are 0.52, 0.44, 0.42 and 0.42 for ASDiv-A, IL, AllArith, and AI2, respectively). In comparison with its CLD across the whole corpus (0.05 in Figure 1), 0.85 is much higher. It is thus suspected that the MWPs within its test/training-set might be very similar to each other. As explained in Footnote #7, a test-set with very similar MWPs hinders the assessment of the true performance of an MWP solver. We thus further compare the CLDs within the test set for a few corpora. Figure 7 illustrates that the CLD=0.27 of MathQA measured within the test set is actually low in comparison with the corresponding CLDs of other corpora (e.g., the means are 0.57 for ASDiv-A and ASDiv corpora).